\documentclass{article} 
\usepackage{nips11submit_e,times}
\usepackage{graphicx}
\usepackage{amsmath}
\usepackage{amsfonts}

\title{
Information theoretic learning of robust deep representations
}

\author{
Nicolas Pinchaud\\
University Pierre et Marie Curie\\
UPMC LIP6\\
Paris, France\\
\texttt{nicolas.pinchaud@gmail.com} \\
}

\nipsfinalcopy 

\begin{document}

\maketitle

\begin{abstract}

We propose a novel objective function for learning robust deep representations of data based on information theory. Data is projected into a feature-vector space such that the mutual information of all subsets of features relative to the supervising signal is maximized.  This objective function gives rise to robust representations by conserving available information relative to supervision in the face of noisy or unavailable features. Although the objective function is not directly tractable, we are able to derive a surrogate objective function. Minimizing this surrogate loss encourages features to be non-redundant and conditionally independent relative to the supervising signal. To evaluate the quality of obtained solutions, we have performed a set of preliminary experiments that show promising results.
\end{abstract}
\vspace{-10pt}
\section{Introduction}
\vspace{-5pt}
The classical pre-training process of deep neural networks is done in an unsupervised scheme. It consists of learning a deep nonlinear representation of the data which is then used to initialize a deep supervised feed-forward neural network. This pre-training procedure is usually done by greedily learning and stacking simple learning modules as described in \cite{NIPS2006_739}. It is hypothesized \cite{Bengio-2009} that unsupervised pre-training is useful because the nonlinear representation captures the manifold shape of the input distribution, such that nonlinear variations in the input become linear variations of the representation vector.

We identify a learning module as a model that provides a conditional distribution $P(B|V)$ involving two random vectors $V$ and $B$. $V$ represents the input or visible variables, and $B$ the features or hidden variables. Recently, much research effort have focused on these modules. It has given rise to a large number of models which essentially differ by the kind of information being extracted from $V$ to form features $B$. We identify this information as the mutual information $I(V,B)$. The use of a generative model of $V$ to learn $P(B|V)$ allows to see what kind of information contributes to $I(V,B)$ with the following decomposition : $$H(V)=H(V|B)+I(V,B)$$
where $H(.)$ is the entropy functional. The underlying modeling hypothesis define the information being conveyed by $I(V,B)$
. For example, in factored RBM \cite{ranzato10}, the factors allow, when $B$ is observed, to model some dependencies between components of $V$. These information are \textit{set in} $H(V|B)$. The remaining information is \textit{put in} $I(V,B)$, for example, this includes higher order dependencies.  
It is also possible to \textit{hide} or \textit{reveal} some information by a pre-processing step of data. Learning a generative model on this transformed data can be easier. For example, a popular pre-processing step is sphering, it corresponds to decorrelate the components of $V$. This helps learning the model ICA \cite{ICA_book} by determining half of its parameters. It is possible to learn $I(V,B)$ without using a generative model of $V$, for example using auto-encoders \cite{Vincent08extractingand}. We believe that a desirable property is to have a mutual information $I(V,B)$ that represents useful information to solve the supervised problem easily, e.g. with a linear model using $B$ as input.

The information $I(V,B)$ can be revealed by more or less complex interactions between components of $B$, this also influences the ease of solving the supervised problem by using the representation $B$. For example, suppose that $V$ is Bernoulli with $P(V=1)=0.5$, and suppose that $B\in \{0,1\}^2$ with an uniform distribution. A generative model of $V$ could be $V=XOR(B_0,B_1)$. In such case, if one component $B_0$ or $B_1$ is not observed, then it is not possible to determine the value of $V$. In other words we have $I(V,B_0)=I(V,B_1)=0~bit$. Information about $V$ is revealed by observation of both components of $B$ : its value is determined by an interaction between components corresponding to the xor function. To minimize interactions between components, one can consider a learning objective that would maximize $I(V,B_i)$ for each components. A particular setting is obtained when components of $B$ are independent and conditionally independent relative to $V$, in this case we have $I(V,B)=\sum_i I(V,B_i)$. We consider a more general objective which consists of maximizing mutual information $I(V,\mathcal{I})$ where $\mathcal{I}$ represents any subset of components of $B$. With an empathize for small subsets, this maximization would lead to representations that are robust : even if some component of $B$ are not observed, we can still have information about $V$. 
 In this work we show that sparse coding \cite{overcomplete}\cite{sparseDBN}\cite{ranzato-nips-07} helps to get such representations. We shall see that we can derive this objective from another one which integrates the supervised signal.

A poor number of models have focused on an explicit integration of the signal of supervision during the pre-training process. This is an important question since there is nothing to guaranty that information $I(V,B)$ is represented on $B$ in such a way that supervision can be easily disentangled by a simple model using $B$ as input. For example, the manifold of the data learned in an unsupervised scheme doesn't guaranty that supervision, e.g. discrete classes, splits the manifold in easily separable parts. Another motivation is that distribution of $V$ may be too complicated to be properly modeled with a simple model. The mutual information $I(V,B)$ that can be learned is limited by the model's capacity. It is then important that this capacity is spent for useful information relative to the supervised task. We denote $Y$ as the variable representing the supervision, e.g. labels. Previous related works \cite{NIPS2006_739}\cite{DRBM}\cite{ranzatosemi} can be interpreted as a joint optimization of $I(V,B)$ and $I(Y,B)$. We propose to maximize mutual information $I(Y,\mathcal{I})$ for any subset $\mathcal{I}$ of component of $B$. This objective leads to distributions $P(Y|B)$ that are robust, because if some component of $B$ are noisy or give misleading information about $Y$, then other components can still fill the gap of information about $Y$. Moreover, we can show that it helps to model $P(Y|B)$ with a simple model like Naive Bayes, because it generates components of $B$ that are conditionally independent relative to $Y$.


\vspace{-10pt}
\section{Learning objective}

\vspace{-5pt}
\subsection{Framework}

We aim to learn a model of $P(Y|X)$, where $X$ and $Y$ are two random vectors, which respectively represent the input and the output, and we have a set $\mathcal{D}$ of samples of their joint distribution. We model $P(Y|X)$ with a deep feed-forward neural network initialized with a deep representation. We hypothesize that the deep representation is a distribution $P(B^{(L)}|X,\theta)$ that factorizes multiple layers as following : 
\vspace{-0.1pt}
\[P(B^{(L)},B^{(L-1)},...,B^{(1)}|X,\theta)=P(B^{(L)}|B^{(L-1)},\theta) \times...\times P(B^{(2)}|B^{(1)},\theta)P(B^{(1)}|X,\theta)\]As in \cite{NIPS2006_739}, $P(B^{(L)}|X,\theta)$ is trained by greedily stacking simpler learning modules that extracts features of previous layer. We model a module by a parameterized distribution $P(B|V,\theta)$, where $V$ and $B$ are observed and hidden random vectors. We have $B=B^{(l)}$ and, if $l=1$, then $V=X$, else $V=B^{(l-1)}$. Note that observations of $V$ also come with observations of $Y$ during the training phase.

We suppose that $B \in \mathbb{R}^m$. We note $B_i$, the $i^{th}$ component of $B$. We note $H$ to designate the Shannon entropy or the differential entropy if variables are continuous, we also note $I$ to designate the mutual information between two variables. 

We suppose that the inference of $B$ is easy by assuming that :
\begin{equation}\label{inde}
P(B|V,\theta) = \prod_i P(B_i|V,\theta)
\end{equation}
\vspace{-20pt}
\subsection{Learning objective}
\vspace{-5pt}
We note $\Pi_n$ the set of all subsets of components of $B$ of size $n$. We consider the following objective :
\vspace{-8pt}
\begin{equation}\label{obj0}
\theta^* = \mbox{arg}\max_{\theta}\limits\sum_{n=1}^{m} \nu_n \sum_{\mathcal{I}\in \Pi_n}\limits I(Y,\mathcal{I}|\theta)
\end{equation}
The coefficients $\nu_n$ are positive hyper-parameters. This objective maximizes mutual information $I(Y,\mathcal{I}|\theta)$, maximization for subsets $\mathcal{I}$ of size $n$ can be emphasized by a high value $\nu_n$. Let $\mathcal{I} \in \Pi_{n+1}$, then for any component $B_i$ in $\mathcal{I}$, we can write $I(Y,\mathcal{I})=I(Y,B_i|\mathcal{J})+ I(Y,\mathcal{J})$, with $\mathcal{J}=\mathcal{I}\backslash \{B_i\}\in \Pi_n$, this consideration allows us to equivalently express (\ref{obj0}) as component-wise sums :
\vspace{-6pt}
\begin{equation}\label{obj1}
\theta^* = \mbox{arg}\max_{\theta}\limits\sum_{n=0}^{m-1} \lambda_n \sum_{(\mathcal{I},B_i)\in \Lambda_n}\limits I(Y,B_i|\mathcal{I},\theta)
\end{equation}

with $\lambda_n=\frac{1}{(n+1)C_{n+1}^m}\sum_{k=n+1}^{m}C_{k}^m\nu_k$,\footnote{the notation $C_k^n$ designates the combination $\dbinom{n}{k}$. The relation can be proved by recursion.} and $\Lambda_n$ the set of pairs, defined by $\Lambda_n = \{(\mathcal{I},B_i) :\mathcal{I} \in \Pi_n, B_i \notin \mathcal{I} \}$.

Let note $\beta_n = \sum_{(\mathcal{I},B_i)\in \Lambda_n}\limits I(Y,B_i|\mathcal{I},\theta)$. To compute $\beta_n$ using the model $P(B|V,\theta)$, we need the distribution of $V$ which is unknown, however we can use the data set $\mathcal{D}$ to get an estimation $\hat{\beta}_n$. Using $\mathcal{D}$, the learning objective becomes :
\vspace{-8pt}
 \begin{equation}\label{obj2}
\theta^* = \mbox{arg}\max_{\theta}\limits\sum_{n=0}^{m-1} \lambda_n \hat{\beta}_n
\end{equation}

We use this objective to learn modules, we expect that stacking them allows to greedily get better solutions with higher values of $\sum_{n=0}^{m-1} \lambda_n \beta_n$.



\vspace{-5pt}
\subsection{Adding constraints to the objective}
\vspace{-5pt}
Since $I(Y,B_i|\mathcal{I},\theta)=H(B_i|\mathcal{I},\theta) - H(B_i|Y,\mathcal{I},\theta)$, the maximization of $\hat{\beta}_n$ can be made by maximizing differences between estimates $H(B_i|\mathcal{I},\theta,\mathcal{D})$ and $H(B_i|Y,\mathcal{I},\theta,\mathcal{D})$. This possibly yield different solutions depending on the value of $H(B_i|\mathcal{I},\theta,\mathcal{D})$. We make a prior hypothesizing that estimate $I(Y,B_i|\mathcal{I},\theta,\mathcal{D})$ is more robust if $H(B_i|\mathcal{I},\theta,\mathcal{D})$ is high with redundant information about $Y$. We suppose that $V$ represents a source of information that may help to satisfy this property. Therefore, we propose to increase the entropy $H(B_i|\mathcal{I},\theta,\mathcal{D})$ by increasing mutual information $I(V,B_i|\mathcal{I},\theta,\mathcal{D})$.
\paragraph{Constrained learning objective :}

We propose to find a solution of (\ref{obj2}) by solving :
%
%
\begin{equation}\label{solve}
\mbox{arg}\max_{\theta}\limits \left(\sum_n  \sum_{(\mathcal{I},B_i)\in \Lambda_n}\limits [\mu_nI(B_i,V|\mathcal{I},\theta,\mathcal{D}) -  \gamma_n H(B_i|Y,\mathcal{I},\theta,\mathcal{D})] \right)
\end{equation}

This learning objective corresponds to a constrained version of (\ref{obj2}) where coefficients $(\mu_n,\gamma_n)$ imply a corresponding coefficient $\lambda_n$.\footnote{Note that this requires some hypothesis if variables are continuous because then we consider the differential entropies which can diverge to $-\infty$. This divergence is avoided if we assume noise in the distribution $P(B|V,\theta)$ which bounds its differential entropy with a finite value. As seen in \cite{Bell95aninformation-maximization}, if the model $P(B|V,\theta)$ is such that $H(B|V,\theta)$ does not depend on $\theta$, maximizing $I(B,V|\theta)$ is equivalent to maximizing $H(B|\theta)$.}

However, solving the problem (\ref{solve}) is not tractable because :
\begin{itemize}
 \item the size of set $\Lambda_n$ is the combination $\dbinom{m}{n+1}$, the sum over its elements is not tractable for $1<n<m-1$,
 \item computing $I(B_i,V|\mathcal{I},\theta,\mathcal{D})$ and $H(B_i|Y,\mathcal{I},\theta,\mathcal{D})$ has a complexity of $O(N^{n+1})$, where $N$ is the number of values that can take a component $B_i$ with a discretization using $log(N)$ bits, and with $n$ the size of $\mathcal{I}$.
\end{itemize}

To find solutions to the problem (\ref{solve}), we propose to consider two sub problems :
\begin{itemize}
 \item maximization of $\sum_n\limits \mu_n \sum_{(\mathcal{I},B_i)\in \Lambda_n}\limits I(B_i,V|\mathcal{I},\theta,\mathcal{D})$,
 \item minimization of $ \sum_n\limits \gamma_n \sum_{(\mathcal{I},B_i)\in \Lambda_n}\limits H(B_i|Y,\mathcal{I},\theta,\mathcal{D})$
\end{itemize}

In the next section we show how we can approximately optimize these functions.
\vspace{-5pt}
\subsection{Maximization of the conditional mutual information}
\vspace{-5pt}
For clarity, we do not write $\theta$ and $\mathcal{D}$ in this section. We show how to maximize :
\begin{equation}\label{maxi}
 \sum_n\limits \mu_n \sum_{(\mathcal{I},B_i)\in \Lambda_n}\limits I(B_i,V|\mathcal{I})
\end{equation}
The idea is as follow :
\begin{itemize}
 \item We learn an mutual information $I(V,B)$ with a classical unsupervised model such as an RBM \cite{EFH}\cite{DBN} or an auto-encoder \cite{Vincent08extractingand}\cite{NIPS2006_739}.
 \item For small sets $\mathcal{I}$, we \textit{spread} the conditional mutual information over components of $B$ to get a lower bound of $I(B_i,V|\mathcal{I})$. This lower bound is a function of $I(V,B)$, thus increasing the mutual information, also increases the conditional mutual information. 
 \item For large sets $\mathcal{I}$, we show how we can use sparsity to increase $I(B_i,V|\mathcal{I})$. 
\end{itemize}

\paragraph{Spreading the information :}

Let say that the information is spread to a depth $n$ if :
\vspace{-4pt}
\begin{equation}\label{spread}
 \forall k : 0\leq k \leq n, \quad \exists c_k,\quad  \forall (\mathcal{I},B_i)\in \Lambda_k,\quad I(B_i,V|\mathcal{I})=c_k
\end{equation}

Spreading information is not tractable for large depth $n$, in practice we wont be able to spread the information for depth $n>1$. Note also that by (\ref{inde}), we have : \begin{equation}
\forall k:0\leq k < n, \quad c_k\geq c_{k+1}                                                                                                                                                                                     \end{equation}

\paragraph{Lower bound :}
If the information is spread to depth $n$, then we have, for $0\leq k \leq n$ : 
\begin{equation}
c_k \geq \frac{I(B,V) - \sum_{0\leq i < k}\limits c_i}{m-k}
\label{borne}
\end{equation} 

\textit{proof :}

\textit{Since the mutual information $I(B,V)$ can be written :}
\begin{equation}\label{decomp}
I(B,V) = I(B_0,V) + I(B_1,V|B_0) + I(B_2,V|B_1,B_0) + ... +  I(B_{m-1},V|B_{m-2},...,B_0)
\end{equation}
\textit{the spreading to depth $n$ implies :}
\[
I(B,V) = \sum_{0\leq i \leq n}\limits c_i + I(B_{n+1},X|B_n,...,B_0) + ... +  I(B_{m-1},X|B_{m-2},...,B_0)
\]
\textit{let} $0\leq k \leq n$,
\textit{the hypothesis (\ref{inde}) allows us to write, for all} $i$ \textit{so that} $0\leq i \leq m-k$ :
\begin{equation}\label{ineg}
I(B_{k+i},V|B_{k-1+i},...,B_{k-1},...,B_0)\leq c_k
\end{equation}
\textit{then we have $I(B,V) \leq \sum_{0\leq i < k}\limits c_i + (m-k)c_k$, which gives (\ref{borne}).}

$\square$

We see that $c_0\geq \frac{I(B,V)}{m}$, therefore $c_0$ can be increased by increasing $I(B,V)$. But this is not the case for $c_k$ with $k>0$. For example, we have $c_1\geq \frac{I(B,V)-c_0}{m-1}$, the bound depends on $c_0$, it is low if $c_0$ is high. However, we can use sparsity to control the growth of $c_0$ and guaranty a higher bound on $c_k$.

\paragraph{Spreading and sparsity :}

Let suppose that information is spread at least to depth zero. If $c_0=\frac{I(B,V)}{m}$, then the decomposition (\ref{decomp}) and the inequality (\ref{ineg}) implies that :
\begin{equation}
 \forall n, \quad  \forall (\mathcal{I},B_i)\in \Lambda_n,\quad I(B_i,V|\mathcal{I})=\frac{I(B,V)}{m}
\end{equation}

Constraining $c_0=\frac{I(B,V)}{m}$ can be done by reducing the entropies $H(B_i)$, we can do this by constraining the probabilities (or densities) $P(B_i=0)$ to high value, this leads $B$ to be sparse. 

Let suppose that information is spread at depth $1$, since we have $c_0\geq c_1$ , constraining $c_0=\frac{I(B,V)}{m}$, does not allow $c_1$ to have higher value than $\frac{I(B,V)}{m}$. Thus sparsity may be useful to increase conditional mutual information for large set $\mathcal{I}$, but too much sparsity may hurt the conditional mutual information for smaller set $\mathcal{I}$.

\vspace{-5pt}
\subsubsection{Optimization in the binary case}
\vspace{-5pt}
We propose to optimize the conditional mutual information in the case where $B \in \{0,1\}^m$. This case allows to easily optimize the spread of information to depth $1$. First we show that in some conditions, we can estimate mutual information by the entropies $H(B_i|\mathcal{I})$, then we deduce a simple way to optimize the spread to depth $1$.

\paragraph{Estimating the mutual information :} If $H(B|V)=0$, then for all $n$, and for all pairs $(\mathcal{I},B_i) \in \Lambda_n$, we have :
\[
I(B_i,V|\mathcal{I}) = H(B_i|\mathcal{I})
\]
This is because $I(B_i,V|\mathcal{I}) = H(B_i|\mathcal{I}) - H(B_i|V,\mathcal{I})$, with $H(B_i|V,\mathcal{I})=0$ which is implied by $H(B|V)=0$.

Optimization of $H(B|V)=0$ may be done by saturating probabilities $P(B_i=1|V)$ to one or to zero. Note that previous work \cite{ICML2011Rifai_455} advocate such optimization, but with motivation related to an invariance property of the representation.

\paragraph{Spreading information to depth one :} 

If we estimate the conditional mutual information by the conditional entropy. And if without loss of generality, we suppose that for all component $P(B_i=1)\leq 0.5$, it can be easily verified that information is spread to depth one if :
\begin{equation}\label{binary_spread}
\begin{array}{ll}
 \exists p_1, \forall i, &P(B_i=1)=p_1\\
 \exists p_{11}, \forall i, \forall j : i\neq j,\quad &P(B_i=1,B_j=1)=p_{11}
 \end{array}
\end{equation}

So under the condition $H(B|V)=0$, spreading information to depth one in the binary case can be done by constraining the probabilities of activation of components to one value, and by constraining the probabilities of joint activation for pairs of components to another value.

Optimization of (\ref{binary_spread}) can be done by minimization of sum of Kullback-Leibler divergences, reintroducing the parameters $\theta$ of the model $P(B|V,\theta)$ :

 \[
     \begin{array}{l}
        d(\theta,p_1) = \sum_{i}\limits D_{KL}(\mathcal{B}(p_1)\|P(B_i|\theta))\\
 
        d^{11}(\theta,p_{11})= \sum_{i,j;i \neq j}\limits D_{KL}(\mathcal{B}(p_{11})\|P(B_iB_j|\theta)) 
     \end{array}
  \]

where $\mathcal{B}(p)$ is the Bernoulli distribution of parameter $p$. Constraints (\ref{binary_spread}) are satisfied if both functions $d(\theta,p_1)$ and $d^{11}(\theta,p_{11})$ equal zero.

We propose to set $p_1$ and $p_{11}$ as hyper-parameters, we can see that $p_1$ controls also the sparsity level of $B$. For simplicity we will choose $p_{11}=p^2_1$, this leads components to be pair-wise independent.

\paragraph{Optimized function :}
Let note $\mathcal{L}_I(\theta, \mathcal{D})$ the loss function optimized to learn mutual information $I(B,V)$ with the help of training set $\mathcal{D}$, and inducing a model $P(B|V,\theta)$. For example $\mathcal{L}_I(\theta,\mathcal{D})$ can refer to the negative log-likelihood of a generative model, or a reconstruction error if $I(V,B)$ is learned using an auto-encoder. We define a loss function which allows us to optimize (\ref{maxi}), by jointly optimizing $I(B,V)$ and spread of information :

\begin{equation}
 \mathcal{L}_{V}(\theta,\mathcal{D}) = \mathcal{L}_I(\theta,\mathcal{D}) + \eta_0d(\theta,p_{1};\mathcal{D}) + \eta_1d^{11}(\theta,p_{11};\mathcal{D})
\end{equation}

Note that this loss does not include explicitly an optimization of $H(B|V)=0$. This is because we hypothesize that the combination of sparsity and model behind $\mathcal{L}_I(\theta,\mathcal{D})$, do not allow do have an high entropy $H(B|V)$. However, relaxing the sparsity constraint and adding an optimization of $H(B|V)=0$ is a path that would be interesting to explore.
\vspace{-5pt}
\subsection{Minimization of the conditional entropy}
\vspace{-5pt}
In this section, we show how to minimize the entropy conditioned with the supervision variable $Y$. This is easier than increasing the mutual information, as seen in previous sub-section, because it consists basically to delete information. We want to minimize :

\begin{equation}\label{mini}
 \sum_n\limits \gamma_n \sum_{(\mathcal{I},B_i)\in \Lambda_n}\limits H(B_i|Y,\mathcal{I},\theta,\mathcal{D})
\end{equation}

Since we have $H(B_i|Y,\mathcal{I},\theta,\mathcal{D})\leq H(B_i|Y,\theta,\mathcal{D})$, we can minimize (\ref{mini}) by minimizing :

\begin{equation}\label{mini2}
 (\max_{i}\limits \gamma_i) \sum_n\limits  H(B_n|Y,\theta,\mathcal{D})
\end{equation}

Minimizing (\ref{mini2}), also minimizes the (conditional) total correlation :

\[
 \sum_n\limits  H(B_n|Y,\theta,\mathcal{D}) - H(B|Y,\theta,\mathcal{D}) 
\]
The (conditional) total correlation is positive and equals zero if and only if components of $B$ are independent conditionally to $Y$. This kind of independence is the hypothesis made by Naive Bayes models, therefore minimization of (\ref{mini2}) allows to use them to model $P(Y|B)$. Particularly, we can prove that, if component of $B$ are independent conditionally to $Y$, if $Y$ is countable, $B$ binary, and that $H(Y|B)=0$
, then $P(Y|B)$ can be modeled by a linear model.


As in the previous sub-section, we suppose that $B$ is binary. Since a lot of supervised problems are classification tasks which involve an uniform discrete random variable $Y$ taking a small number of values, we propose to develop an optimizable function to minimize (\ref{mini2}) under this hypothesis.
\vspace{-5pt}
\subsubsection{Optimization in the binary case and small number of classes}
\vspace{-5pt}
We suppose that $B\in \{0,1\}^m$, and that $Y$ is a uniform discrete random variable taking values in $\mathcal{Y}=\{y_0,...,y_{K-1}\}$, we have $P(Y=y)=\frac{1}{K}$. We also suppose that $m$ is sufficiently large and $K$ sufficiently small, so that for all component $B_i$, we have $P(B_i=1) \leq \frac{1}{K}$. Let $B_i$ be a component of $B$, and $Q$ be a joint distribution of $(B_i,Y)$, we have :
\[
    \begin{array}{l}
       P(B_i=1) \leq \frac{1}{K} \\
       P(B_i)=\sum_y Q(B_i,Y=y) \\
       P(Y=y) = \frac{1}{K} = Q(B_i=0,y)+Q(B_i=1,y)
    \end{array}    
 \]
 We can show that a distribution $Q$ that satisfies hypothesis, minimizes $H(B_i|Y;Q)$ if it also satisfies :
 \begin{equation}\label{h_min}
  \begin{array}{ll}
       \exists y_i \in \mathcal{Y}, \forall y\neq y_i, & Q(B_i=1|y)=0    
    \end{array}  
 \end{equation}
We assign to each component $B_i$ a class by defining a surjection\footnote{This is possible if $m\geq K$, we suppose that this is the case.} $\phi:\{0,...,m-1\} \rightarrow \mathcal{Y}$.
The distribution $P(B_i,Y|\theta)$ is of the form (\ref{h_min}), if and only if we have :
\[
\forall y\neq \phi(i), \forall v: P(V=v,Y=y)\neq 0,\quad P(B_i=1|v,y,\theta)=0 
\]
This can be optimized using data set $\mathcal{D}$ by minimizing the following Kullback-Leibler divergences :
\begin{equation}
 \mathcal{L}_Y(\theta,\mathcal{D}) = \sum_{(x,y)\in \mathcal{D}}\sum_n \textbf{1}_{y\neq\phi(n)} D_{KL}(\mathcal{B}(0) \| P(B_n|v_x,y,\theta,\mathcal{D}))
\end{equation}
where $\mathcal{B}(0)$ is the Bernoulli distribution of parameter $p=0$, $v_x$ is a sample of $P(V|X=x,\theta)$, and $\textbf{1}_{y\neq\phi(n)}$ is the indicator function, it equals $1$ if $y\neq\phi(n)$, equals $0$ otherwise.
\vspace{-5pt}
\subsection{Joint optimization}
\vspace{-5pt}
We propose a global loss that jointly optimizes the maximization of conditional mutual information and minimization of conditional entropies.
\vspace{-5pt}
\begin{equation}
 \mathcal{L}(\theta,\mathcal{D}) = \mathcal{L}_V(\theta,\mathcal{D}) + \eta_y \mathcal{L}_Y(\theta,\mathcal{D})
\end{equation}
The hyper parameters are $\{\eta_0,\eta_1,p_1,\eta_y\}$ (with $p_{11}=p_1^2$).
\vspace{-10pt}
\section{Experiments}
\vspace{-5pt}
We used two data-sets, Mnist and Cifar-BW. Mnist is the well known data set of digit classification problem. We used 50000 examples for training, 10000 examples for validation, and we tested on the official 10000 examples. Cifar-BW is a gray-scale version of Cifar-10 data set \cite{cifar}, obtained by averaging the RGB values. This data set represents a image-classification task with 10 classes. We trained on 40000 examples, 10000 for validation, and 10000 for test.

%

\begin{figure}[!h]
 \centering
 \begin{tabular}{cc}
      \includegraphics[width=4cm]{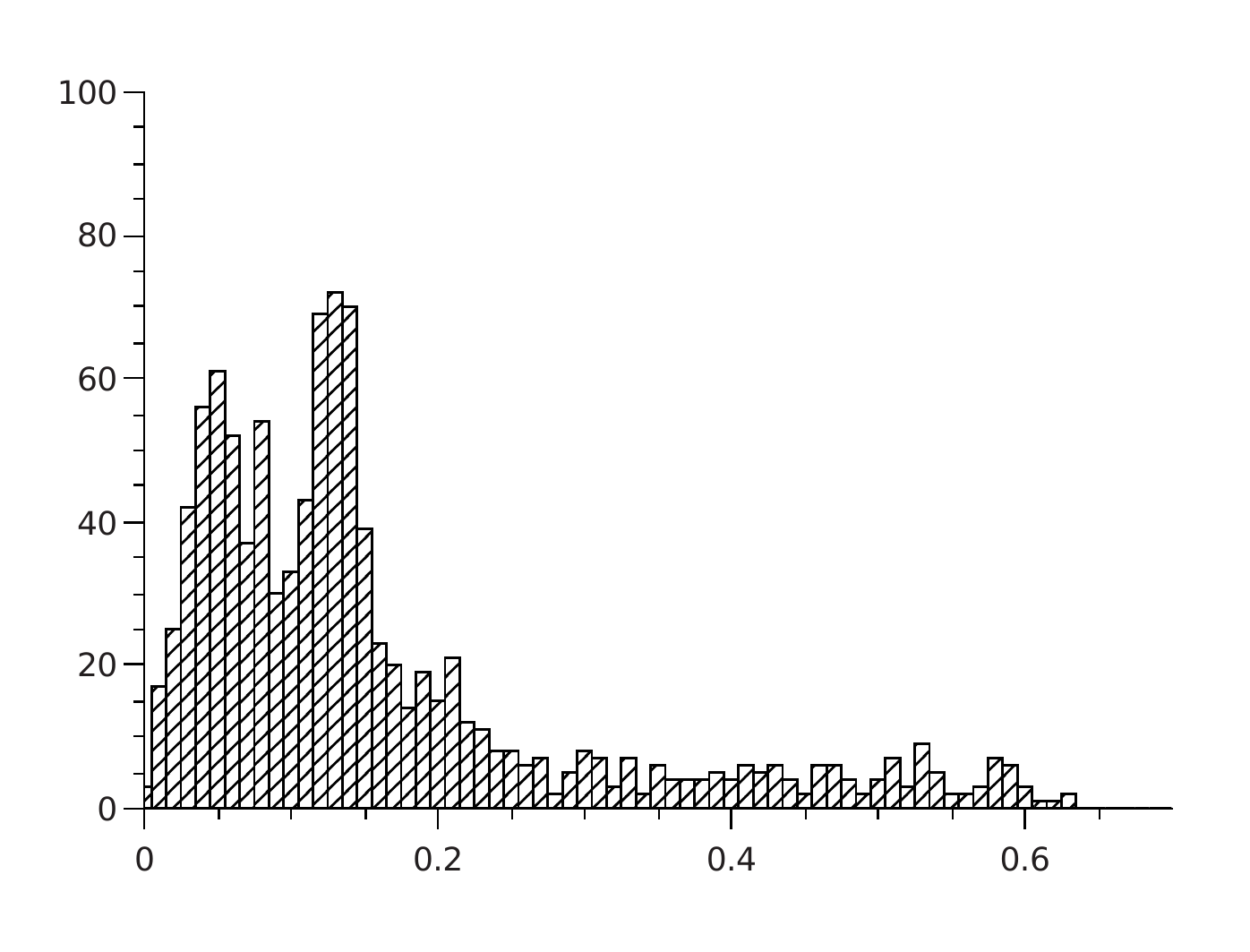} & \includegraphics[width=4cm]{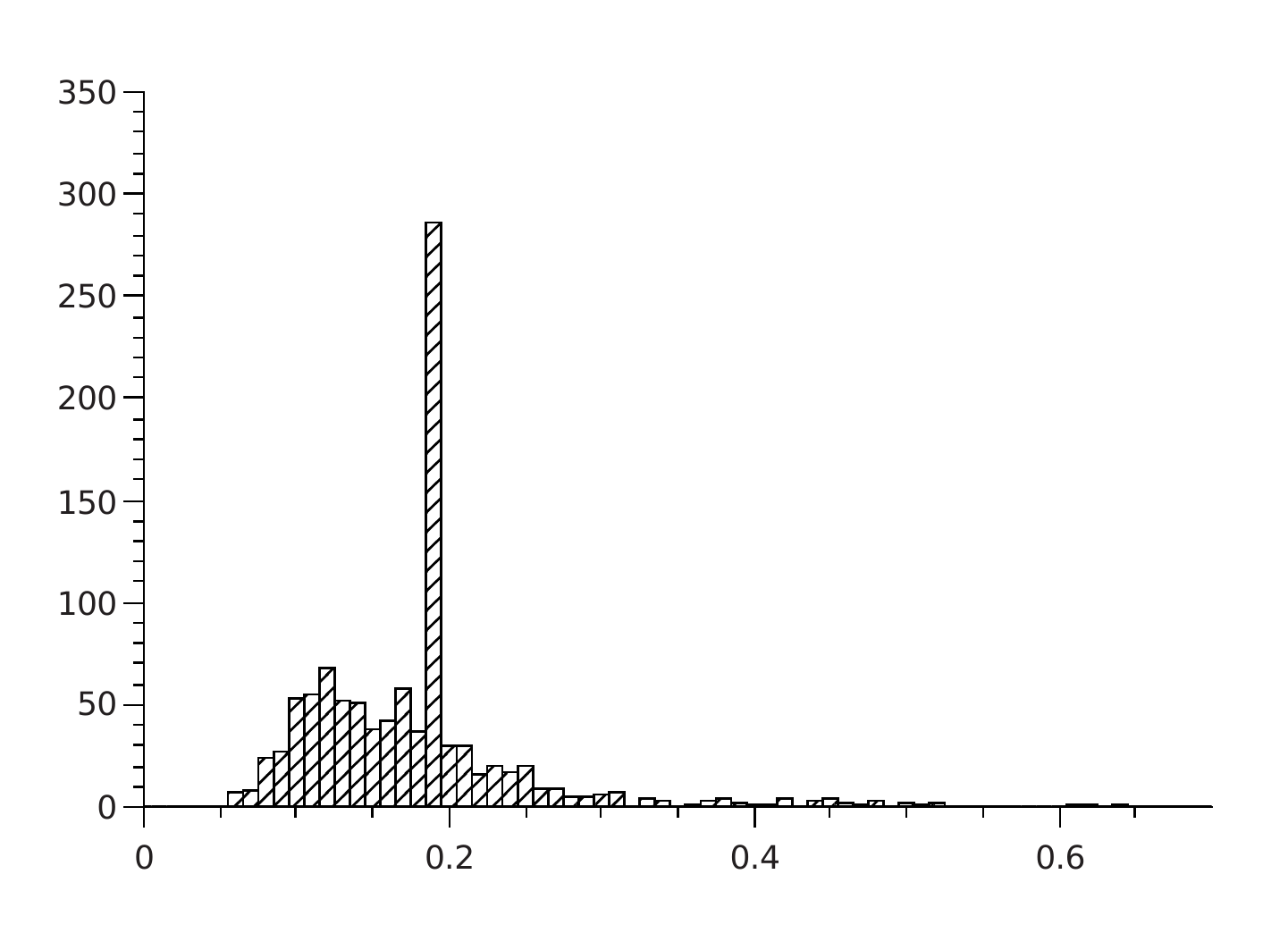} \\
    Sparse RBM & $\mathcal{L}_V$  
   \end{tabular}
 \caption{Histograms of minimal information (in nat) owned by each components with a conditioning to another component, for Sparse RBM on the left, by optimizing spread to depth one on the right.}
 \label{utilities_hist}
\end{figure}

We have trained a one hidden layer representation on Mnist with a RBM and we have optimized sparsity and spread of information to depth one (we note $L_V$). We have compared the conditional mutual information obtained with those obtained with a Sparse RBM \cite{sparseDBN}. Sparse RBM is a RBM trained with sparsity regularization which corresponds to the optimization of spread to depth zero. Figure \ref{utilities_hist} shows the histogram of minimal mutual information owned by components with a conditioning set of size one, for a component $B_n$, it is $\min_{i \neq n}\limits I(X,B_n|B_i)$, where $X$ follows the distribution of Mnist. We see that Sparse RBM does not avoid two components to be completely redundant, this is characterized by an information of 0~nat, while spreading the information over components to depth one, prevents this discrepancy.

The figure \ref{exp1} compares classification performances on Mnist and Cifar-BW when spreading is optimized to depth zero only (Sparse RBM/GRBM), or to depth one ($L_V$). For the loss $\mathcal{L}_I$ we used an RBM on Mnist, or an Gaussian RBM (GRBM) \cite{EFH}\cite{cifar} on Cifar-BW. The Sparse GRBM is a GRBM trained with the same regularization as Sparse RBM. For each case we show the best performance obtained after a grid search on hyper-parameters. We fixed $\eta_0=\eta_1$, we used a model with one hidden layer with 1024 components. Each result represents the mean classification error on 30 runs using different random initialization of parameters.

\begin{figure}[!h]
 \centering
 
\begin{tabular}{|c|c|c|}\hline
 & Mnist & Cifar-BW\\ \hline
Sparse RBM / GRBM & 1.36 & 49.1 \\ \hline
$L_V$ & 1.31 & 48.6\\ \hline

\end{tabular}
\caption{Classification error obtained on Mnist and Cifar-BW, the results are obtained with 30 runs on Mnist, and 5 runs on Cifar-BW. The model had one hidden layer with 1024 components. The pre-training module used to optimize $\mathcal{L}_V$ was an RBM for Mnist and a Gaussian RBM (GRBM) for Cifar-BW.}
\label{exp1}
\end{figure}

\begin{figure}[!h]
 \centering
 \includegraphics[width=6cm]{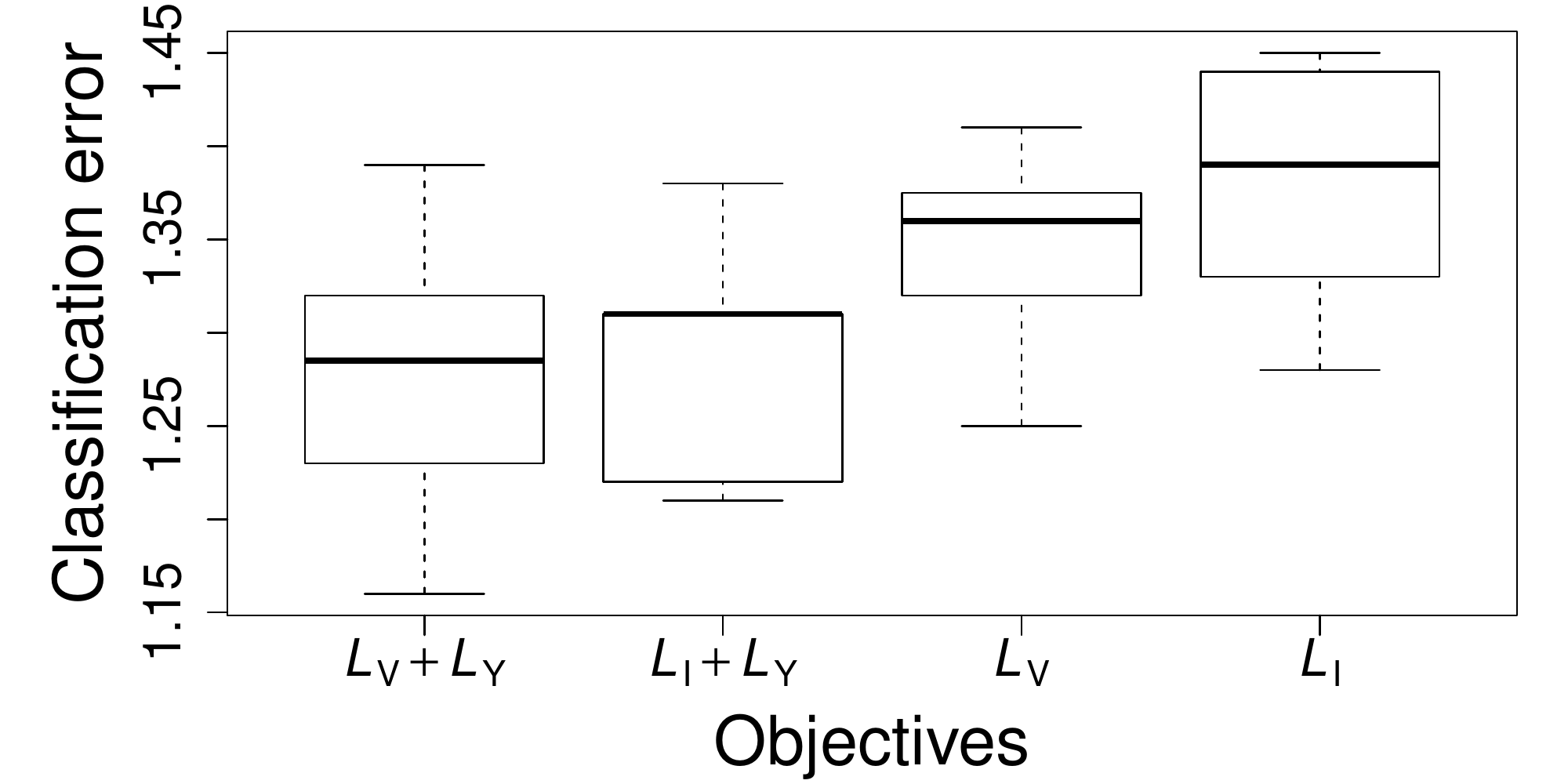}
 \caption{Classification error on Mnist, with a RBM alone ($L_I$), optimizing spread to depth one ($L_V$), optimizing a RBM along with $\mathcal{L}_Y$ ($L_I+L_Y$), and optimizing both $\mathcal{L}_V$ and $\mathcal{L}_Y$. The model has two hidden layers with 1000 components.}
\label{exp2}
\end{figure}

The figure \ref{exp2} shows box plot of classification error on Mnist using a model with 2 hidden layers with 1000 components each, with optimizing a RBM only ($L_I$), optimizing information spread to depth one ($L_V$), optimizing a RBM along with $\mathcal{L}_Y$ ($L_I+L_Y$), and optimizing both $\mathcal{L}_V$ and $\mathcal{L}_Y$. \footnote{Number of runs are 5 for $L_I+L_Y$, 6 for $L_V$, 11 for $L_I$, and 18 for $L_V + L_Y$.} Optimizing $\mathcal{L}_V$ alone yield worse performance  on a model with 2 hidden layers (we got an error mean of $1.34$), than with one hidden layer only (we got $1.31$, see Figure \ref{exp1}). However effect of $\mathcal{L}_V$ appears to be beneficial when optimized along with $\mathcal{L}_Y$ (we got $1.28$).

\begin{figure}[!h]
 \centering
 \begin{tabular}{ccc}
      \includegraphics[width=2.5cm]{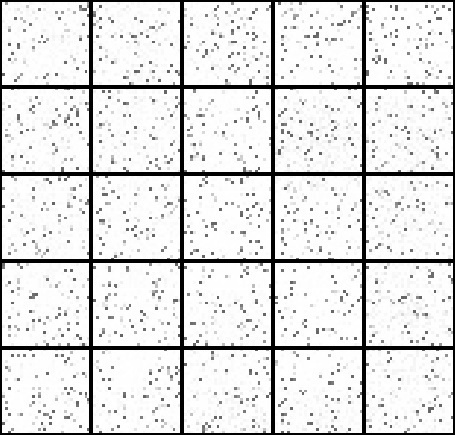} & \includegraphics[width=2.5cm]{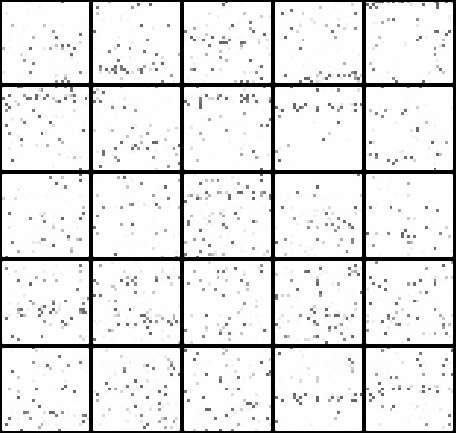} & \includegraphics[width=2.5cm]{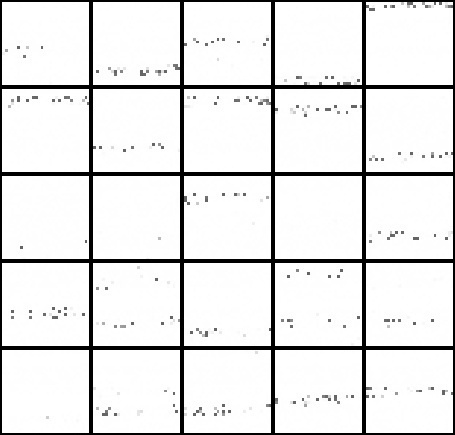} \\
    First hidden layer & Second hidden layer & Third hidden layer
   \end{tabular}
 \caption{Effect of $\mathcal{L}_Y$ on the probability of components. We see that distribution of components shifts to the distribution of the supervision $Y$ as the layer depth increases, while the loss $\mathcal{L}_V$ tries to keep as much information as possible about $X$. See text for details.}
 \label{frange}
\end{figure}

The figure \ref{frange} shows the effect of $\mathcal{L}_Y$ on probability of components. Each square displays the hidden representation of an example from Mnist, the black dots represent components with high probability to be equal to one, components on the same row are specialized for the same label (their $\phi$ value are equal). For each layer, we have increased the parameter $\eta_Y$ by a factor $100$. We see that distribution of components shifts to the distribution of the supervision $Y$ as the depth increases, while the loss $\mathcal{L}_V$ tries to keep as much information as possible about $X$.
\vspace{-20pt}
\section{Discussion}
\vspace{-5pt}
We have introduced an objective based on information theory that aims to discover robust representations. The objective is not tractable, therefore we have derived a surrogate loss. This loss is a weighted sum of two terms $\mathcal{L}_V$ and $\mathcal{L}_Y$. The first one, $\mathcal{L}_V$, maximizes the entropy of components of representation while their redundancy is kept small. We have seen relations to sparse coding methods. We have proposed to increase the entropy using information expressed by the input distribution, this links our approach with unsupervised pre-training methods. The second term, $\mathcal{L}_Y$, minimizes entropy of components conditioned to the supervising signal. This leads to components that are conditionally independent relative to the supervision. This allows distribution of the supervision to be modeled with a Naive Bayes model using the representation as input. We have proposed to work in the context of deep learning, where deep representation is obtained by greedily training and stacking simple modules. The final model is a deep feed-forward neural network initialized with the representation. A set of experiments have shown promising results. We have seen that pre-training neural network with $\mathcal{L}_V$ alone gives good results for shallow representation, but addition of one hidden layer worsens performance. However, using both losses $\mathcal{L}_V$ and $\mathcal{L}_Y$ gives our best performance. This advocates the integration of supervised signal during pre-training. Although, more experiments have to be done to confirm experimental results.


\newpage

\bibliographystyle{plain}
\bibliography{workshop_nips.bib}

\newpage

\end{document}